\documentclass[10pt,twocolumn,letterpaper]{article}

\usepackage{cvpr}
\usepackage{times}
\usepackage{epsfig}
\usepackage{graphicx}
\usepackage{amsmath}
\usepackage{amssymb}
\usepackage{float}
\usepackage{subcaption}
\usepackage{multirow}

\usepackage[pagebackref=true,breaklinks=true,letterpaper=true,colorlinks,bookmarks=false]{hyperref}

\cvprfinalcopy 

\def\cvprPaperID{8} 

\begin{document}

\title{Prediction-Tracking-Segmentation}
\author{Jianren Wang\hspace{0.5cm}Yihui He\hspace{0.5cm}Xiaobo Wang\hspace{0.5cm}Xinjia Yu\hspace{0.5cm}Xia Chen\\
Carnegie Mellon University\\
{\tt\small \{jianrenw,he2,xiaobow,xinjiay,xiac\}@andrew.cmu.edu}
}

\twocolumn[{%
\maketitle
    \begin{center}
    \includegraphics[width=\textwidth]{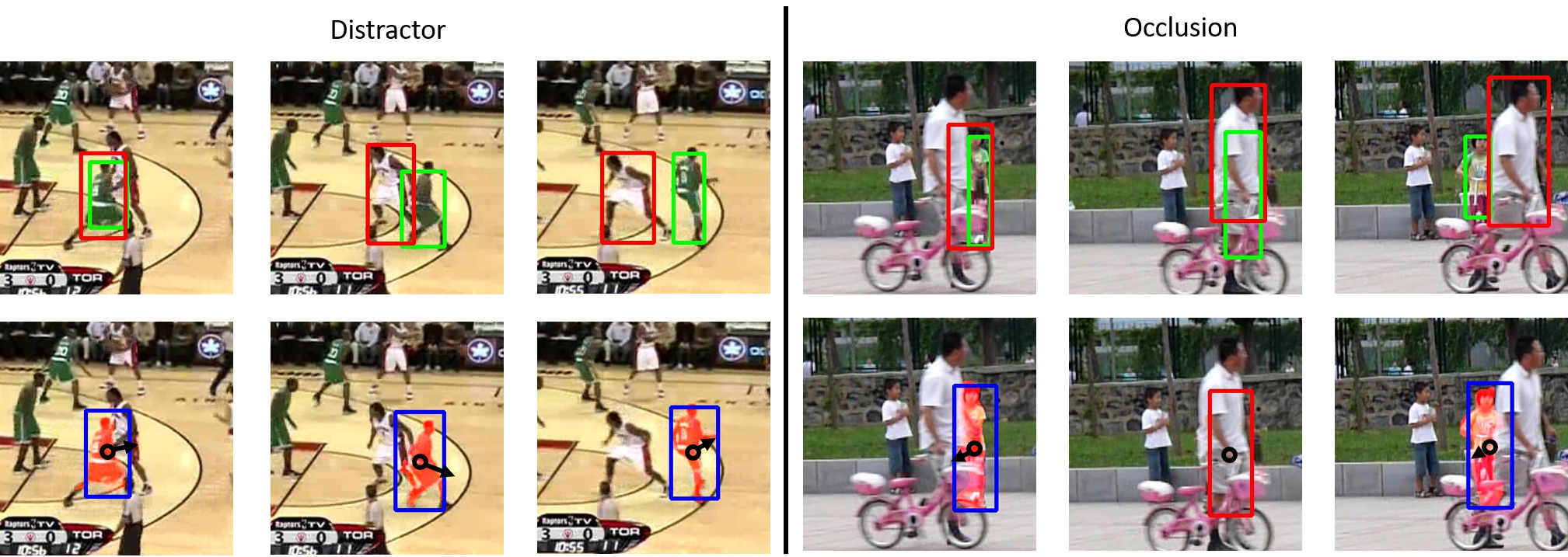}\\
  \captionof{figure}{We propose a prediction driven method for tracking and segmentation in videos. The first row shows the results of state-of-the-art tracker SiamMask~\cite{wang2018fast} (red) and ground truth (green). The second row shows the motion prediction (black arrow) and tracking results (with the blue bounding box and red segmentation mask) of our Prediction-Tracking-Segmentation (PTS) model. Our method improves the robustness against distractors and occlusions. (better view with color)}
    \label{fig:teaser}
     \end{center}
}]

\begin{abstract}
We introduce a prediction driven method for visual tracking and segmentation in videos. Instead of solely relying on matching with appearance cues for tracking, we build a predictive model which guides finding more accurate tracking regions efficiently. With the proposed prediction mechanism, we improve the model robustness against distractions and occlusions during tracking. We demonstrate significant improvements over state-of-the-art methods not only on visual tracking tasks (VOT 2016 and VOT 2018) but also on video segmentation datasets (DAVIS 2016 and DAVIS 2017).

\end{abstract}


\section{Introduction}

A human can track, segment or interact with fast moving objects with surprising accuracy~\cite{smeets1998visuomotor}, even in the cases the objects are under deformations, occlusions and illumination changes~\cite{wu2015object}. What is the key component in human perception to make this happen?

In fact, tracking and segmenting moving object appears at a very early stage of human perception. Even 4-month-old infants can track moving objects with his or her eyes and reaching for them~\cite{von1982eye}. A professional athlete can even interact with objects at breakneck speed. (\eg, a baseball player can hit a 100-mph baseball). To do so, the human brain has to overcome its delays in neuronal transmission through prediction~\cite{nijhawan1994motion, hogendoorn2018predictive, cavanagh2013flash} and saves times for processing what we see by using only local information~\cite{van2018motion,cassanello2008neuronal}. Besides prediction, researchers further point out that humans have multiple temporal scales for tracking objects with different speeds~\cite{holcombe2009seeing}.




Inspired by these observations from human perception, in this paper, we propose a two-stage tracking method driven by prediction. Given a tracking result in time $t$, we first predict the approximate object location in the next frame (in time $t+1$) without seeing it. Based on the prediction results, we then further refine the localization as well as the segmentation results by using the appearance input in time $t+1$. The refined tracking and segmentation results can help us back to update a better prediction model, which will be applied again in the successive frames. 

Concretely, in the first stage of prediction, we use an extrapolation method to estimate the object position in the future frame (in time $t+1$) and simulate the multiple temporal scales effect in human perception through adaptive search region in the same frame. With prediction driven tracking and segmentation, we name our method Prediction-Tracking-Segmentation (PTS).


Our approach offers several unique advantages. First, through prediction of object position, we free tracking from using only appearance information. As most tracking and segmentation methods can only discriminate foreground from the non-semantic background~\cite{zhu2018distractor}, the performance suffers significantly when the target object is surrounded by similar objects (know as distractors~\cite{zhu2018distractor}). Prediction also improves the robustness of our model against occlusions. Note that occlusions largely prevent most appearance-based methods from extracting useful information. We show in the experiments that our method improves the tracking performance by a large margin under both cases. We visualize part of the results in Fig.~\ref{fig:teaser}.

Second, through the usage of adaptive search region around the predicted area, PTS significantly decreases the information required to process and thus has a large potential to increase the inference speed. To achieve this, we propose to focus on smaller local regions when objects have slower speeds and smaller sizes, and vice versa. This approach allows better segmentation performance, less missing and identity switching. 


We evaluate our framework all major tracking datasets: VOT-2016~\cite{10.1007/978-3-319-48881-3_54}, VOT-2018~\cite{Kristan2018a}. We demonstrate that our framework achieves state-of-the-art performance, both qualitatively and quantitatively. We also show competitive results against semi-supervised VOS approaches on DAVIS-2016~\cite{perazzi2016benchmark} and DAVIS-2017~\cite{pont20172017}.

To summarize, our main contributions are three-fold: First, inspired by visual cognition theory, we propose PTS to unify predict, tracking and segmentation in a single framework. Second, we propose an adaptive search region module to process information effectively. We indicate that our proposed achieves competitive performance on VOT and VOS datasets.

\section{Related Works}

In this section, we briefly overview three research areas relative to our proposed method.

\paragraph{Video Object Tracking}

In tracking community, significant attention has been paid to discriminative correlation filters (DCF) based methods~\cite{bolme2010visual,ma2015long,li2014scale,galoogahi2017learning}. These methods allow discriminating between the template of an arbitrary target and its 2D translations at a breakneck speed. MOSSE ~\cite{bolme2010visual} is the pioneering work which proposes a fast correlation tracker by minimizing the squared error. Performance of DCF-based trackers has then been notably improved through the using of multi-channel features~\cite{henriques2015high,danelljan2014adaptive,kiani2013multi}, robust scale estimation~\cite{danelljan2014accurate,danelljan2017discriminative}, reducing boundary effects~\cite{danelljan2015learning,kiani2015correlation} and fusing multi-resolution features in the continuous spatial domain~\cite{danelljan2016beyond}.

Tracking through Siamese Network is also an important approach~\cite{koch2015siamese,tao2016siamese,bertinetto2016fully,valmadre2017end}. Instead of learning a discriminative classifier online, the idea is to train a deep siamese similarity function offline on pairs of video frames. At test time, this function is used to search for the candidate most similar to the template given in the starting frame on a new video, once per frame. The pioneering work is SINT~\cite{tao2016siamese}. Similarly, GOTURN~\cite{held2016learning} used deep regression network to predict the motion between successive frames. SiamFC ~\cite{bertinetto2016fully} implemented a fully convolutional network to output the correlation response map with high values at target locations, which set a basic form of modern Siamese framework. Many following works have been proposed to improve the accuracy while maintain fast inference speed by adding semantic branch~\cite{He_2018_CVPR}, using region proposals~\cite{li2018high}, hard negative mining~\cite{zhu2018distractor}, ensembling~\cite{he2018towards}, deeper backbone~\cite{li2018siamrpn++} and high-fidelity object representations~\cite{wang2018fast}.

With the assumption that objects are under minor displacement and size change in consecutive frames, most modern trackers, including all the ones mentioned above, use a steady search region, which is centered on the last estimated position of the target with the same ratio. Although it is very straightforward, this oversimplified prior often fails in occlusion, motion change, size change, camera motion, as it is evident in the examples of Figure 1. This motivated us to propose a tracker able to adaptively set the search region. 
\paragraph{Video Forecasting}

The ability to predict and therefore to anticipate the future is an important attribute of intelligence. Many methods are developed to improve the temporal stability of semantic video segmentation. Luc et al. ~\cite{luc2017predicting} develop an autoregressive convolutional neural network that learns to generate multiple future frames iteratively. Similarly, Walker et al.~\cite{walker2017pose} uses a VAE to model the possible future movements of humans in the pose space. Instead of generating future states directly, many methods attempt to propagate segmentation from preceding input frames~\cite{jin2017video,nilsson2018semantic,jaderberg2015spatial}.

Unlike previous work, inspired by human perception, we extract a motion model for each object and set up a new search region for segmentation according to the motion model. 
\paragraph{Video Object Segmentation}

Video Object Segmentation (VOS) have been divided into three categories based on the level of supervision required: unsupervised, semi-supervised and supervised. We briefly review the VOS focusing on semi-supervised setting, which is usually formulated as a temporal label propagation problem.  To exploit consistency between video frames, many methods propagate the first segmentation mask through temporal adjacent ones~\cite{bao2018cnn,marki2016bilateral,tsai2016video} or even entire video~\cite{jampani2017video,jang2017online}. Another approach is to process video frames independently~\cite{perazzi2017learning} and usually heavily rely on fine-tuning~\cite{caelles2017one}, data augmentation~\cite{khoreva2017lucid} and model adaption~\cite{voigtlaender2017online}.



\section{Method}

\begin{figure*}[ht]
\begin{center}
   \includegraphics[width=0.8\textwidth]{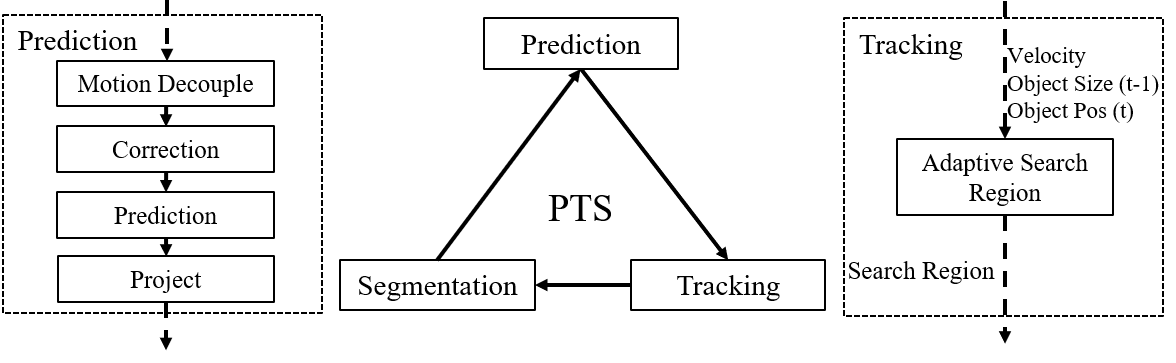}
  \caption{An overview of our method. Our method is composed of prediction part, tracking part and segmentation part.}
  \label{fig:pipeline}   
\end{center}
\end{figure*}

To unify prediction, tracking and segmentation with adaptive search region, our model consists of: (i) prediction module: estimate object position and velocity in an unseen frame (Section~\ref{sec:method:Prediction}) (ii) tracking module: adaptively limit the search region for further processing (Section~\ref{sec:method:tracking}) (iii) segmentation module: a fully-convolutional Siamese framework to segment foreground object from given search region (Section~\ref{sec:method:Segmentation}). We show our framework in Figure~\ref{fig:pipeline}.

\subsection{Prediction Module}\label{sec:method:Prediction}

\begin{figure}[t!]
\begin{center}
   \includegraphics[width=0.4\textwidth]{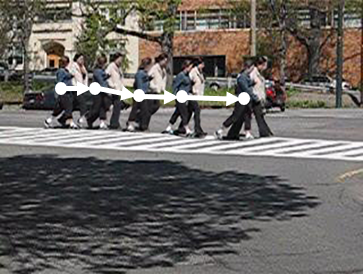}
  \caption{One example for decoupling background motion and mapping object motion to reference frame (arrows illustrates the movement of object center)}
  \label{fig:motion}   
\end{center}
\end{figure}

In video object tracking and segmentation, most methods do not consider the time consistency of object motion. In other words, most methods predict a   zero-velocity-object and thus set a local search region centered on the last estimated position of the target~\cite{bertinetto2016fully,li2018high,zhu2018distractor,wang2018fast}.

However, these methods only consider appearance features of the current frame, and hardly benefit from motion information. This leads to great difficulty in distinguishing between instances that look like the template, known as distractors~\cite{zhu2018distractor} or under occlusion, fast motion and camera motion. To solve this problem, our proposed tracker takes full advantage of the motion information. 

Object motion in a given image is the superposition of camera motion and object motion. The former is random while the latter should satisfy Newton's First Law~\cite{newton1999principia}. We first pick a reference frame ($F_{r_k}$, $r_k$ denotes $k^{th}$ reference frame) every $n$ frames and thus separate the long video into several pieces of short n-frame videos.

Second, we adopt the method proposed by ARIT~\cite{6751553} to decouple the camera motion and object motion within each short video. ARIT assumes that pending detection frame ($F_{r_k+t}$) and its reference frame ($F_{r_k}$) are related by a homography ($H_{r_k,r_k+t}$). This assumption holds in most cases as the global motion between neighboring frames is usually small. To estimate the homography, the first step is to find the correspondences between two frames. As mentioned in ARIT, we combine SURF features~\cite{willems2008efficient} and motion vectors from the optical flow to generate sufficient and complementary candidate matches, which is shown to be robust~\cite{gauglitz2011evaluation,6751553}. Here we use PWCNet~\cite{sun2018pwc} for dense flow generation.

As a homography matrix contains 8 free variables, at least 4 background points pairs should be used. We calculate the least square solution of eq.~\ref{eq:7} and optimize it to obtain robust solution through RANSAC~\cite{fischler1981random}, where $p_{r_k}^{bp}$ and $p_{r_k+t}^{bp}$ denotes random selected background matching pairs in $F_{r_k}$ and $F_{r_k+t}$ using the above mentioned features. Given the assumption that the background occupies more than half of the images, we partition matching points between frames into 4 pieces, and then one point is randomly chosen inside each selected piece to improve the efficiency of RANSAC algorithm. 

\begin{equation} \label{eq:7}
H_{r_k,r_k+t}\times p_{r_k}^{bp}=p_{r_k+t}^{bp}
\end{equation}

For simplicity, all following calculations are under reference coordinate and project back to the new coming frame without further noticing.

Fig.\ref{fig:motion} illustrates the working principle of decoupling step. The origin video for Fig.\ref{fig:motion} is a handheld video with trembling background. The motion of the pedestrians in the origin video is highly unpredictable with huge background uncertainties. However, by mapping the target frame towards the reference frame, the movement for pedestrians could be more predictable and continuous. 

To find the most representative point of object position, we calculate the "center of mass" of object segmentation using eq.~\ref{eq:8}

\begin{equation} \label{eq:8}
P=average(p^{o})
\end{equation}






Random noise from background motion estimation and mask segmentation might be introduced to the object position prediction, which could influence the accuracy of prediction. To achieve a better estimation for object states, we utilize Kalman Filter to provide accurate position information based on the measurements from current and former frames. As a classical tracking algorithm, the Kalman filter estimates the position of the object in two steps: prediction and correction. In the prediction step, it predicts the target state based on the dynamic model (eq.~\ref{predictor}) and generates the search region for the Siamese network to achieve object segmentation. Therefore, the measurement for object position in the next frame could be computed with eq.~\ref{eq:8}. Then, in the correction step, the position measurement would be updated with higher certainty given the position measurement from the Siamese network, which benefits the accuracy of predictions for future frames.

The dynamic model for object position update could be formulated as:
\begin{equation} 
    \label{predictor}
    \hat{x}_{t|t-1} = F_k \hat{x}_{t-1|t-1} + w_t
\end{equation}

In eq.\ref{predictor},  $\hat{x}_{t|t-1}$ is the priori state estimation given observations up to time $t-1$, which is in the form of 4-dimension vector ($[x,y,dx,dy]$) with position information. It is worth to mention that the velocity terms ($dx,dy$) are predicted by using extrapolation between the information from time $t-1$ and $t$. $w_t$ is the random noise existing in the system. And $F_t$ is the transition matrix from time $t-1$ to $t$.

After predicting the states, the Kalman filter uses measurements to correct its prediction during the correction steps using eq.~\ref{update predictor}. In the equation, $\hat{y}_t$ is the residuals between the prediction and measurement. And $K_t$ is the optimal Kalman gain given from the predicted error covariance ($P_{t|t-1}$), measurement matrix ($H_t$) and measurement margin covariance ($S_t$), as shown in eq.~\ref{ Kalman gain}. It is worth to mention that, as Kalman filter is a recursive algorithm, the predicted error covariance (P) should be updated as well based on the estimation results.




\begin{equation}
    \label{update predictor}
    \hat{x}_{t|t} = \hat{x}_{t|t-1} + K_t \hat{y}_t
\end{equation}

\begin{equation}
    \label{ Kalman gain}
    K_t = P_{t|t-1}{H_t}^T{S_t}^T
\end{equation}

The motion consistency between video frames in different sliced videos with different reference frames could be an issue because the initialization of the velocity for the reference frame could be critical to the accuracy of the position update. To maintain the motion consistency, we choose the $n_{th}$ frame, which is the last frame in the sliced video, as the next reference frame with the refined position and velocity estimation from Kalman filter based on the former reference frame. Therefore, the velocity of the object, with respect to the new frame, could be initialized by mapping the refined velocity towards the new reference.

\subsection{Tracking Module}\label{sec:method:tracking}

Inspired by human perception, we dynamically set up a new search region in the coming frame centered at the predicted object position. We project the estimated object center position back to the pending detection frame using eq.~\ref{eq:11}.

\begin{equation} \label{eq:11}
H_{r_k,r_k+t}\times P_{r_k}=P_{r_k+t}
\end{equation}

Given the estimated position, we setup the search region $S$ accordingly using the similar method as in ~\cite{henriques2015high}:

\begin{equation} \label{eq:10}
S=k\sqrt{(w+p)(h+p)}
\end{equation}

\begin{equation}
    \label{parameter}
    k = 1 + 2\times sigmoid(||v||_2- T)
\end{equation}
where $p=\dfrac{w + h}{2}$. To achieve the adaptive search region, the search region size $S$ would be modified with respect to the predicted velocity using eq.\ref{parameter}. In the equation, $v$ is the velocity predicted by Kalman filter and $T$ is the threshold for velocity. The search region is cropped center at $P_{r_k+t}$ on the frame $F_{r+t}$, and then resized in $255 \times 255$.

To make the one-shot segmentation framework suitable for tracking task. We adopt the optimization strategy used for the automatic bounding box generation proposed in VOT-2016~\cite{10.1007/978-3-319-48881-3_54} as it offers the highest IOU and mAP as reported in ~\cite{wang2018fast}.

\subsection{Segmentation Module}\label{sec:method:Segmentation}

We adopt the SiamMask framework~\cite{wang2018fast}, which achieves a good balance between the accuracy and speed. SiamMask propose to use an offline trained fully-convolutional network to simultaneously collect binary segmentation mask, detection bounding box and objectness score. First, the Siamese network compares an template image $z$ ($w \times h \times 3$) against a (larger) search image $x$ to obtain a dense response map $g_{\theta}(z,x)$. The two inputs are processed by the same CNN $f_{\theta}$, yielding two feature maps that are cross-correlated:
\begin{equation} \label{eq:1}
g_{\theta}(z,x)=f_{\theta}(z)\star f_{\theta}(x)
\end{equation}
Each spatial element of the response map $g_{\theta}^n(z,x)$ represent a similarity between the template image $z$ and $n^{th}$ candidate window in $x$. Second, a three-branch head calculates binary segmentation mask, detection bounding box and objectness score, respectively. The mask branch predicts a ($w \times h$) binary mask $m^n$ from each spatial element $g_{\theta}^n(z,x)$. The box branch regresses $k$ bounding boxes from each spatial element $g_{\theta}^n(z,x)$, where $k$ is the number of anchors. And the score branch estimates the corresponding objectness score. 
\begin{equation} \label{eq:2}
m^n=h_\phi(g_{\theta}^n(z,x))
\end{equation}
\begin{equation} \label{eq:3}
b_{i=1,..,k}^n=h_\sigma(g_{\theta}^n(z,x))
\end{equation}
\begin{equation} \label{eq:4}
s_{i=1,..,k}^n=h_\varphi(g_{\theta}^n(z,x))
\end{equation}
A multi-task loss is used to optimise the whole framework. 
\begin{equation} \label{eq:5}
L=\lambda_1 L_{mask}+\lambda_2 L_{box}+\lambda_3 L_{score}
\end{equation}

We refer readers to ~\cite{pinheiro2015learning,pinheiro2016learning} for understanding mask branch and ~\cite{ren2015faster, li2018high} for understanding region proposal branch.



\section{Experiments}
\label{experiments}

In this section, we evaluate our approach on three tasks: motion prediction, visual object tracking (VOT 2016 and VOT 2018)
and semi-supervised video object segmentation (on DAVIS 2016 and DAVIS 2017). It is worth noticing that our method does not depend on the selection of tracking and segmentation methods. To better evaluate the efficiency of our method, we adopt SiamMask~\cite{wang2018fast} with provided pretrained model as our online tracking and segmentation method.

\subsection{Evaluation for motion prediction accuracy}
\paragraph{Datasets and settings}
We adopt two widely used benchmark data set to evaluate the performance the motion prediction: VOT 2016~\cite{10.1007/978-3-319-48881-3_54} and VOT 2018~\cite{Kristan2018a}. Both of them are annotated with the rotated bounding box. Both datasets contain 60 public sequences with different challenging factors: camera motion, object motion change, object size change,  occlusion and illumination change, which makes it extremely challenging for object motion prediction~\cite{Kristan2018a}.
We use eq.~\ref{eq:8} to predict object position and eq.~\ref{predictor} to extrapolate object velocity. For our baseline method, the predicted position of the next frame (t+1) is always the same as the current frame (t), while object velocity is always predicted as 0. The ground truth position is set as the center of the annotated rotated bounding box, while the velocity is the difference between two consecutive positions. We evaluate the position error from ground truth with Euclidean distance and velocity error with Euclidean distance, cosine distance and magnitude distance. Cosine distance is the cosine value between predicted velocity and ground truth velocity (the higher, the better). Magnitude distance is the absolute difference between the absolute value of predicted velocity and ground truth velocity.
We adopt the reinitialize mechanism as used in the official VOT toolkit when the segmentation has no overlap with ground truth. We reinitialize the tracking method with ground truth after five frames.

\paragraph{Results on VOT 2016 and VOT 2018}

\begin{figure}[ht]
\begin{center}
   \includegraphics[width=0.4 \textwidth]{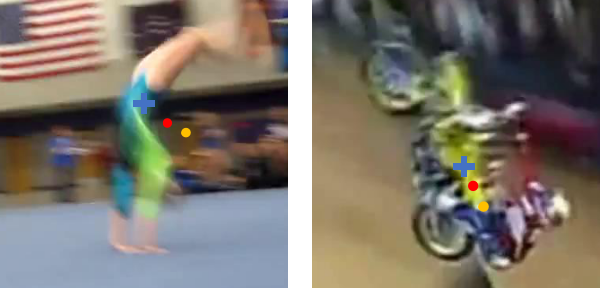}
  \caption{Object center predictions generated by SiamMask and PTS (red for PTS, yellow for SiamMask and blue cross for ground truth) (better view with color)}
  \label{fig:prediction error}   
\end{center}
\end{figure}

\begin{table}[ht!]
\begin{center}
\begin{tabular}{c|c|c}
\hline
Dataset & Tracker &   Pos Err. \\ \hline  
\multirow{2}{*}{VOT 2016} & SiamMask & 16.281 \\ \cline{2-3}
& PTS(ours)       & 8.198 \\ \hline
\multirow{2}{*}{VOT 2018} & SiamMask       & 14.593 \\ \cline{2-3}
& PTS(ours)      & 8.744 \\ \hline

\end{tabular}
\caption{Position prediction error on VOT 2016 and VOT 2018}
\label{prediction error}
\end{center}
\end{table}

\begin{table}[ht!]
\begin{center}
\begin{tabular}{c|c|c|c|c}
\hline
Dataset & Tracker & MSE Err. & Cosine & Mag.\\ \hline  
\multirow{2}{*}{VOT 2016} & SiamMask & 8.274 & - & 8.274\\ \cline{2-5}
& PTS(ours)       &   4.596 & 0.667 & 3.190 \\ \hline
\multirow{2}{*}{VOT 2018} & SiamMask &   7.006 & - & 7.006\\ \cline{2-5}
& PTS(ours)      &    4.298 & 0.793 & 2.929 \\ \hline

\end{tabular}
\caption{Velocity prediction error on VOT 2016 and VOT 2018}
\label{velocity error}
\end{center}
\end{table}

\begin{figure}[ht]
\begin{center}
   \includegraphics[width=0.4 \textwidth]{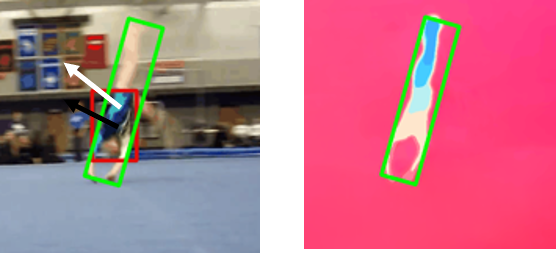}
  \caption{Velocity prediction generated by PTS (white for goundtruth, black for prediction, both extended by 5 times longer for better visualization)(better view with color)}
  \label{fig:velocity error}   
\end{center}
\end{figure}

Table.\ref{prediction error} presents the comparison of position prediction results using SiamMask and PTS based on VOT 2016 and VOT 2018 datasets. 

As it is shown in the table, for both of these two datasets, the PTS method could dramatically reduce the prediction errors of the object position. The mean square error for object position on VOT 2018 could be reduced by half from 16 pixels to 8 pixels. Meanwhile, Fig.\ref{fig:prediction error} shows when the object velocity is high, the PTS method could provide a prediction more accurate compared with the SiamMask, which does not consider the influence of object motion. The results prove that the decoupling strategy could reduce the background uncertainty and the Kalman filter would provide a relatively reliable prediction for object position in the next frame. Higher accuracy for object position prediction could benefit the generation of search regions for object tracking and eventually improve the performance of object segmentation.

For velocity, as can be seen in Table.~\ref{velocity error}, our method significantly reduce the estimation error. In VOT 2018, PTS achieves 0.763 cosine distance, which is about 37 degree divergence from ground truth velocity direction. The main cause of the error is that objects are not always rigid, thus "center of mass" can approximate the overall motion of the object~\ref{fig:velocity error}. The size change of objects will further increase the prediction error. However, with the correction procedure of the Kalman filter, this error (noise) can be stabilized. One possible solution to decrease velocity prediction error is tracking each part of non-rigid objects and grouping all parts together to get the final prediction~\cite{ramanan2007tracking}.

\begin{figure*}[ht]
\begin{center}
   \includegraphics[width=\textwidth]{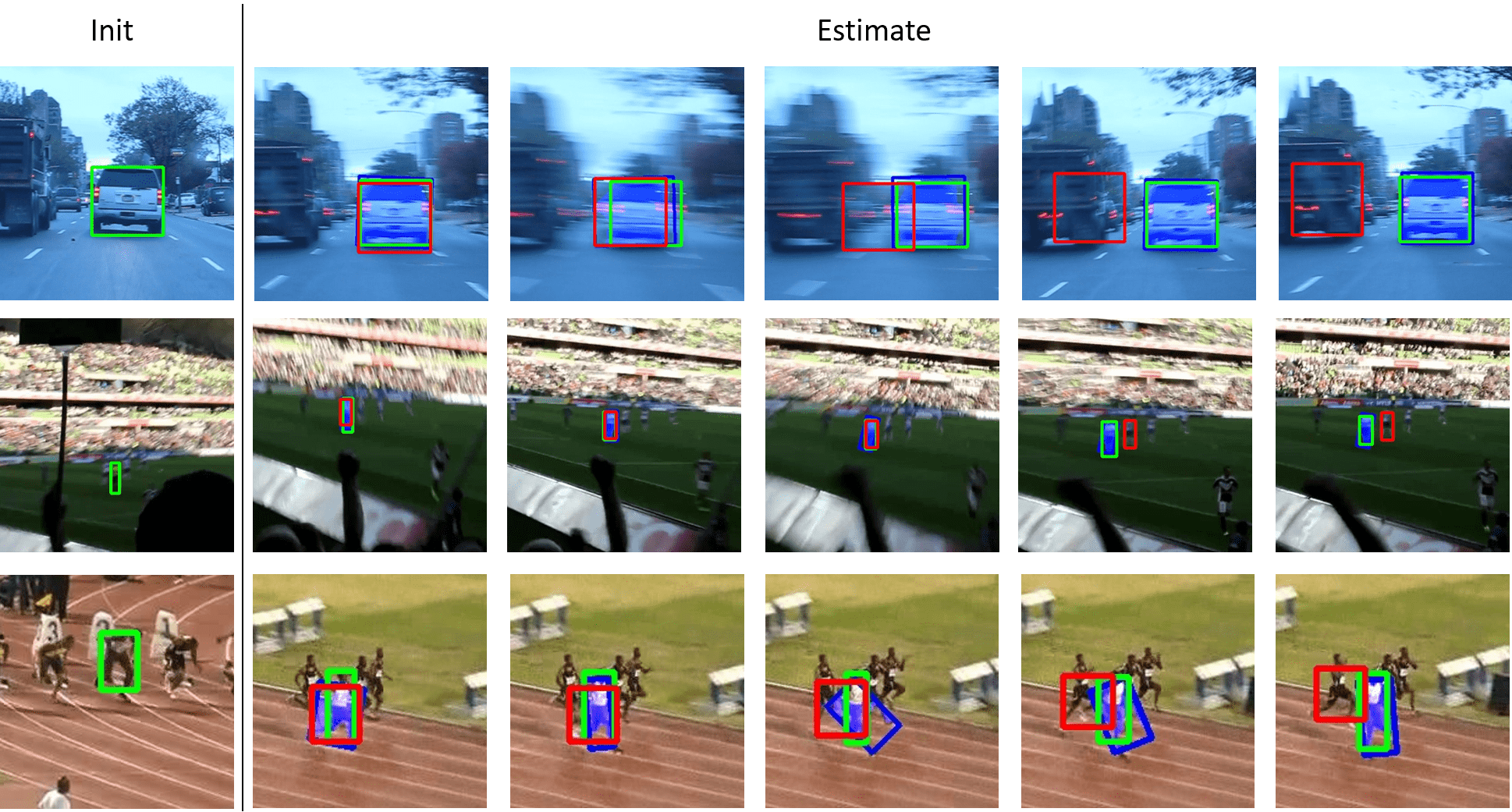}
  \caption{Qualitative result of our method : green box is the ground truth, red box is the bounding box from SiamMask, and blue box is our bounding box for the mask.}
  \label{Qualitative result}   
\end{center}
\end{figure*}

\subsection{Evaluation for visual object tracking}
\paragraph{Datasets and settings}
Similarly, we adopt three widely used benchmarks for the evaluation of the object tracking task: VOT 2016, VOT 2018 and compare against the state-of-the-art using official metric: Expected Average Overlap (EAO), which considers both accuracy and robustness of a tracker~\cite{Kristan2018a}. We use VOT 2018 to conduct an experiment to discuss the performance under different conditions further.  

\paragraph{Results on VOT 2016}

\begin{table}[]
\begin{center}
\begin{tabular}{c|c|c|c}
\hline
           & \multicolumn{3}{c}{VOT 2016} \\ \hline
Trackers   & A        & R       & EAO     \\ \hline
SiamMaks   & 0.639    & 0.214   & 0.433   \\ \hline
PTS (Ours) & 0.642    & 0.144   & 0.471   \\ \hline
\end{tabular}
\end{center}
\caption{Comparison with SiamMask on VOT 2016 }
\label{VOT 2016} 
\end{table}

Table \ref{VOT 2016} present comparisons of tracking performance between PTS and other state-of-the-art models based on VOT 2016 datasets. Our model improves the robustness by 30\%, and provide an 8.8\% gain of EAO, which achieves 0.471.  

\paragraph{Results on VOT 2018}

\begin{table*}[]
\begin{center}
\resizebox{\textwidth}{!}{\begin{tabular}{c|cccccccccccc}
 & DaSiamRPN & SA\_Siam\_R & CPT & DeePTSRCF & DRT & RCO & UPDT & SiamMask & SiamRPN & MFT & LADCF & Ours \\ \hline
EAO ↑ & 0.326 & 0.337 & 0.339 & 0.345 & 0.356 & 0.376 & 0.378 & 0.380 & 0.383 & 0.385 & 0.389 & \textbf{0.397} \\
Accuracy & 0.569 & 0.566 & 0.506 & 0.523 & 0.519 & 0.507 & 0.536 & 0.609 & 0.586 & 0.505 & 0.503 & 0.612 \\
Robustness & 0.337 & 0.258 & 0.239 & 0.215 & 0.201 & 0.155 & 0.184 & 0.276 & 0.276 & 0.140 & 0.159 & 0.220 \\ \hline
\end{tabular}}
\caption{Comparison with the state-of-the-art under EAO, Accuracy, and Robustness on the VOT 2018 benchmark.}
\label{all result} 
\end{center}
\end{table*}

In Table~\ref{all result} we compare our PTS against eleven recently published state-of-the-art trackers on the VOT 2018 benchmark. We establish a new state-of-the-art tracker with 0.397 EAO and 0.612 accuracy. In particular, our accuracy outperforms all existing Correlation Filter-based trackers. This is very easy to understand since our baseline SiamMask relies on deep feature extraction which is much richer than all existing Correlation Filter-based methods. However, PTS even outperforms the baseline SiamMask method, which is very interesting. Previous research shows Siamese based trackers have strong center bias despite the appearances of test targets~\cite{li2018siamrpn++}. Thus, by estimating the center of the search region more accurate, Siamese trackers can also achieve better regression result (\eg, bounding box detection or object segmentation). Besides, PTS achieves the highest robustness among all Siamese based trackers. This is even exhilarating because one of the key vulnerability of Siamese based trackers is the low robustness. The main reason is that most Siamese networks can only discriminate foreground from the non-semantic background~\cite{zhu2018distractor}. In other words, Siamese based trackers are not appearance sensitive enough and always suffer from distinguishing surrounding objects. Our proposed PTS adopts a straightforward strategy and shows huge improvement of robustness from 0.276 to 0.220, which provides another strategy to achieve better robustness: by setting more accurate and targeted search region.

To further analysis where the improvements come from, we show the qualitative results of PTS and our baseline\ref{Qualitative result}. Just as mentioned above, the robustness comes from less tracking object switching and missing. For example, as for the car scenario in figure \ref{Qualitative result}, when the camera shakes, the center of the search region of SiamMask will shift to the left of the tracking car, and finally catches the truck. On the contrary, since our model considers camera motion, the center of our search region stays on the tracking car. This stability comes from the decoupling of camera motion. Another example is Bolt, the third row in figure \ref{Qualitative result}. When Bolt accelerates, SiamMask will be easily distracted by other runners, but our PTS model won't fail because it considers the speed of Bolt. This stability comes from object velocity estimation. These unique features greatly help the performance of PTS under large camera motion, fast object motion and occlusion. Simply speaking, by predicting object position accurately, we can focus on more targeted search region and thus achieve better detection and segmentation performance. 

\begin{table}[]
\begin{center}
\begin{tabular}{c|c|c|c}
Datasets & Methods & J & F \\ \hline
\multirow{2}{*}{DAVIS 2016} & SiamMask & 0.713 & 0.674 \\  \cline{2-4}
& PTS(ours) & 0.732 & 0.692 \\ \hline
\multirow{2}{*}{DAVIS 2017} & SiamMask & 0.543 & 0.585 \\  \cline{2-4}
& PTS(ours) & 0.554 & 0.604 \\  \hline
\end{tabular}
\caption{J and F Results on DAVIS 2016 and DAVIS 2017 }
\label{Davis} 
\end{center}
\end{table}

Table~\ref{Davis}  presents the comparison of vos results using SiamMask and PTS based on DAVIS 2016 and DAVIS 2017 datasets.

\begin{figure*}[ht]
\begin{center}
   \includegraphics[width=\textwidth]{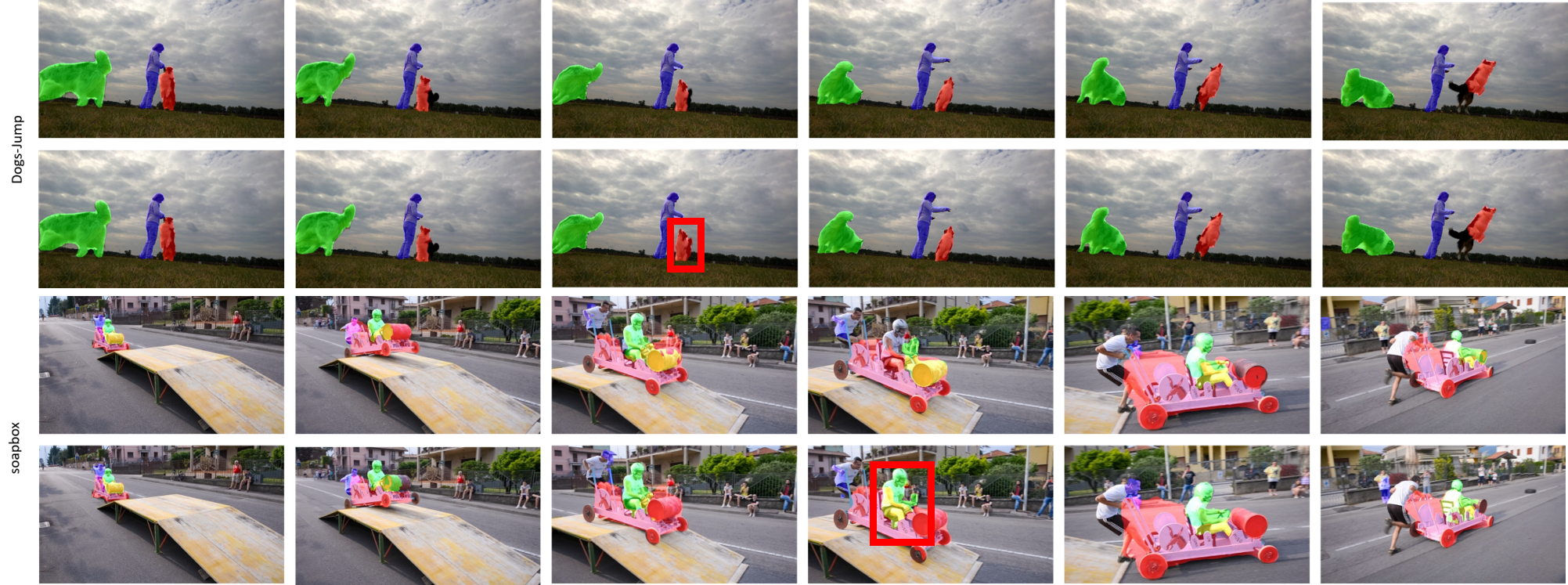}
  \caption{Qualitative result of SiamMask and PTS on DAVIS: First row and third row are the results from SiamMask. Second row and fourth row are the results from same videos using PTS. (better view with color)}
  \label{fig:davis}   
\end{center}
\end{figure*}
\subsection{Evaluation for video object segmentation}

\paragraph{Datasets and settings}

We report the performance of PWT on standard VOS datasets DAVIS 2016~\cite{perazzi2016benchmark} and DAVIS 2017~\cite{pont20172017}. For both datasets, we use the official performance measures: the Jaccard index (J) to express region similarity and the F-measure (F) to express contour accuracy. Since we use SiamMask as our baseline, we adopt the semi-supervise setup. We fit bounding boxes to object masks in the first frame and use these bounding boxes to initialize our proposed PTS.  

\paragraph{Results on DAVIS 2016 and DAVIS 2017}

The effect of our approach is limited on DAVIS 2016 and DAVIS 2017 datasets. The main reason is that DAVIS datasets have less camera motion or fast object motion, which are the major gain from our method. However, segmentation can still benefit from more accurately cropped search region. \eg, The dog in the third frame of "Dogs-Jump" video is segmented more completely through PTS. However, SiamMask misses the tail of the same dog during segmentation. Another example is the person in the fourth frame of "Soap-Box" video. PTS separates this person from the soapbox, however, SiamMask mixes its segmentation with the surrounding pixels. Further, SiamMask fails to distinguish the person mask from the drum of the soapbox because the drum occupies the previous position of the person, which can not be handled without motion assumption. Though our pre-tracking procedure, PTS can separate specific instance from its neighboring instance and thus get a more accurate segmentation. We show that our proposed PTS has very large potential especially under segmentation of crowded scenarios. 
For more qualitative results, please refer to Fig~\ref{fig:davis}. 

\subsection{Ablation studies}

Table~\ref{ablation} compares the influence of each module in our model. Based on VOT 2018 dataset, we evaluate the influence of tracking and prediction module and compare their performance with the baseline approach (SiamMask) and PTS. It can be observed from Table~\ref{ablation} that the prediction module which uses Kalman filter to update the position of objects plays an important role in PTS that most EAO improvements seem to be introduced by the prediction module. Moreover, for the tracking module, which is the adaptive search region update module, the influence is a little bit limited with only 0.02 EAO increase. However, as we can see from Table~\ref{ablation}, both of these two modules have the potential to improve accuracy.

\begin{table}[]
\begin{tabular}{c|ccc}
 & EAO & Accuracy & Robustness \\ \hline
SiamMask & 0.380 & 0.609 & 0.276 \\
SiamMask + Tracking & 0.382 & 0.610 & 0.268 \\
SiamMask + Prediction & 0.394 & 0.611 & 0.234 \\
PTS & 0.397 & 0.612 & 0.220 \\ \hline
\end{tabular}
\caption{Ablation studies for Tracking and Prediction modules on VOT 2018 dataset. }
\label{ablation} 
\end{table}

\section{Conclusion}

In conclusion, we introduce a prediction driven method for visual tracking  and  segmentation  in  videos. Instead  of  solely  relying  on  matching  with  appearance  cues  for  tracking,  we build a predictive model which provides guidance on finding more accurate tracking regions efficiently. We show that this simple idea significantly improve the robustness in VOT and VOS challenges and achieve state-of-the-art performance in both tasks. We hope our work can inspire more studies in considering the relationship among three main challenges in video understanding: prediction, tracking and segmentation. 

\nocite{he2019addressnet,he2018vehicle,zhu2019feature,he2018amc,liang2017single,he2018softer,he2017channel}
\nocite{chen2018integration,wang2018vertical,wang2017prioritization,xia2017validation}
{\small
\bibliographystyle{ieee}
\bibliography{egbib}

\begin{thebibliography}{10}\itemsep=-1pt

\bibitem{bao2018cnn}
Linchao Bao, Baoyuan Wu, and Wei Liu.
\newblock Cnn in mrf: Video object segmentation via inference in a cnn-based
  higher-order spatio-temporal mrf.
\newblock In {\em Proceedings of the IEEE Conference on Computer Vision and
  Pattern Recognition}, pages 5977--5986, 2018.

\bibitem{bertinetto2016fully}
Luca Bertinetto, Jack Valmadre, Joao~F Henriques, Andrea Vedaldi, and Philip~HS
  Torr.
\newblock Fully-convolutional siamese networks for object tracking.
\newblock In {\em European conference on computer vision}, pages 850--865.
  Springer, 2016.

\bibitem{bolme2010visual}
David~S Bolme, J~Ross Beveridge, Bruce~A Draper, and Yui~Man Lui.
\newblock Visual object tracking using adaptive correlation filters.
\newblock In {\em Computer Vision and Pattern Recognition (CVPR), 2010 IEEE
  Conference on}, pages 2544--2550. IEEE, 2010.

\bibitem{caelles2017one}
Sergi Caelles, Kevis-Kokitsi Maninis, Jordi Pont-Tuset, Laura Leal-Taix{\'e},
  Daniel Cremers, and Luc Van~Gool.
\newblock One-shot video object segmentation.
\newblock In {\em Proceedings of the IEEE conference on computer vision and
  pattern recognition}, pages 221--230, 2017.

\bibitem{cassanello2008neuronal}
Carlos~R Cassanello, Abhay~T Nihalani, and Vincent~P Ferrera.
\newblock Neuronal responses to moving targets in monkey frontal eye fields.
\newblock {\em Journal of neurophysiology}, 2008.

\bibitem{cavanagh2013flash}
Patrick Cavanagh and Stuart Anstis.
\newblock The flash grab effect.
\newblock {\em Vision Research}, 91:8--20, 2013.

\bibitem{chen2018integration}
Shuyang Chen, Jianren Wang, and Peter Kazanzides.
\newblock Integration of a low-cost three-axis sensor for robot force control.
\newblock In {\em 2018 Second IEEE International Conference on Robotic
  Computing (IRC)}, pages 246--249. IEEE, 2018.

\bibitem{danelljan2014accurate}
Martin Danelljan, Gustav H{\"a}ger, Fahad Khan, and Michael Felsberg.
\newblock Accurate scale estimation for robust visual tracking.
\newblock In {\em British Machine Vision Conference, Nottingham, September 1-5,
  2014}. BMVA Press, 2014.

\bibitem{danelljan2017discriminative}
Martin Danelljan, Gustav H{\"a}ger, Fahad~Shahbaz Khan, and Michael Felsberg.
\newblock Discriminative scale space tracking.
\newblock {\em IEEE transactions on pattern analysis and machine intelligence},
  39(8):1561--1575, 2017.

\bibitem{danelljan2015learning}
Martin Danelljan, Gustav Hager, Fahad Shahbaz~Khan, and Michael Felsberg.
\newblock Learning spatially regularized correlation filters for visual
  tracking.
\newblock In {\em Proceedings of the IEEE International Conference on Computer
  Vision}, pages 4310--4318, 2015.

\bibitem{danelljan2016beyond}
Martin Danelljan, Andreas Robinson, Fahad~Shahbaz Khan, and Michael Felsberg.
\newblock Beyond correlation filters: Learning continuous convolution operators
  for visual tracking.
\newblock In {\em European Conference on Computer Vision}, pages 472--488.
  Springer, 2016.

\bibitem{danelljan2014adaptive}
Martin Danelljan, Fahad Shahbaz~Khan, Michael Felsberg, and Joost Van~de
  Weijer.
\newblock Adaptive color attributes for real-time visual tracking.
\newblock In {\em Proceedings of the IEEE Conference on Computer Vision and
  Pattern Recognition}, pages 1090--1097, 2014.

\bibitem{fischler1981random}
Martin~A Fischler and Robert~C Bolles.
\newblock Random sample consensus: a paradigm for model fitting with
  applications to image analysis and automated cartography.
\newblock {\em Communications of the ACM}, 24(6):381--395, 1981.

\bibitem{galoogahi2017learning}
Hamed~Kiani Galoogahi, Ashton Fagg, and Simon Lucey.
\newblock Learning background-aware correlation filters for visual tracking.
\newblock In {\em ICCV}, pages 1144--1152, 2017.

\bibitem{gauglitz2011evaluation}
Steffen Gauglitz, Tobias H{\"o}llerer, and Matthew Turk.
\newblock Evaluation of interest point detectors and feature descriptors for
  visual tracking.
\newblock {\em International journal of computer vision}, 94(3):335, 2011.

\bibitem{he2018towards}
Anfeng He, Chong Luo, Xinmei Tian, and Wenjun Zeng.
\newblock Towards a better match in siamese network based visual object
  tracker.
\newblock In {\em European Conference on Computer Vision}, pages 132--147.
  Springer, 2018.

\bibitem{He_2018_CVPR}
Anfeng He, Chong Luo, Xinmei Tian, and Wenjun Zeng.
\newblock A twofold siamese network for real-time object tracking.
\newblock In {\em The IEEE Conference on Computer Vision and Pattern
  Recognition (CVPR)}, June 2018.

\bibitem{he2018amc}
Yihui He, Ji Lin, Zhijian Liu, Hanrui Wang, Li-Jia Li, and Song Han.
\newblock Amc: Automl for model compression and acceleration on mobile devices.
\newblock In {\em Proceedings of the European Conference on Computer Vision
  (ECCV)}, pages 784--800, 2018.

\bibitem{he2019addressnet}
Yihui He, Xianggen Liu, Huasong Zhong, and Yuchun Ma.
\newblock Addressnet: Shift-based primitives for efficient convolutional neural
  networks.
\newblock In {\em 2019 IEEE Winter Conference on Applications of Computer
  Vision (WACV)}, pages 1213--1222. IEEE, 2019.

\bibitem{he2018vehicle}
Yihui He, Xiaobo Ma, Xiapu Luo, Jianfeng Li, Mengchen Zhao, Bo An, and Xiaohong
  Guan.
\newblock Vehicle traffic driven camera placement for better metropolis
  security surveillance.
\newblock {\em IEEE Intelligent Systems}, 33(4):49--61, 2018.

\bibitem{he2018softer}
Yihui He, Xiangyu Zhang, Marios Savvides, and Kris Kitani.
\newblock Softer-nms: Rethinking bounding box regression for accurate object
  detection.
\newblock {\em arXiv preprint arXiv:1809.08545}, 2018.

\bibitem{he2017channel}
Yihui He, Xiangyu Zhang, and Jian Sun.
\newblock Channel pruning for accelerating very deep neural networks.
\newblock In {\em Proceedings of the IEEE International Conference on Computer
  Vision}, pages 1389--1397, 2017.

\bibitem{held2016learning}
David Held, Sebastian Thrun, and Silvio Savarese.
\newblock Learning to track at 100 fps with deep regression networks.
\newblock In {\em European Conference on Computer Vision}, pages 749--765.
  Springer, 2016.

\bibitem{henriques2015high}
Jo{\~a}o~F Henriques, Rui Caseiro, Pedro Martins, and Jorge Batista.
\newblock High-speed tracking with kernelized correlation filters.
\newblock {\em IEEE Transactions on Pattern Analysis and Machine Intelligence},
  37(3):583--596, 2015.

\bibitem{hogendoorn2018predictive}
Hinze Hogendoorn and Anthony~N Burkitt.
\newblock Predictive coding of visual object position ahead of moving objects
  revealed by time-resolved eeg decoding.
\newblock {\em Neuroimage}, 171:55--61, 2018.

\bibitem{holcombe2009seeing}
Alex~O Holcombe.
\newblock Seeing slow and seeing fast: two limits on perception.
\newblock {\em Trends in cognitive sciences}, 13(5):216--221, 2009.

\bibitem{jaderberg2015spatial}
Max Jaderberg, Karen Simonyan, Andrew Zisserman, et~al.
\newblock Spatial transformer networks.
\newblock In {\em Advances in neural information processing systems}, pages
  2017--2025, 2015.

\bibitem{jampani2017video}
Varun Jampani, Raghudeep Gadde, and Peter~V Gehler.
\newblock Video propagation networks.
\newblock In {\em Proceedings of the IEEE Conference on Computer Vision and
  Pattern Recognition}, pages 451--461, 2017.

\bibitem{jang2017online}
Won-Dong Jang and Chang-Su Kim.
\newblock Online video object segmentation via convolutional trident network.
\newblock In {\em Proceedings of the IEEE Conference on Computer Vision and
  Pattern Recognition}, pages 5849--5858, 2017.

\bibitem{jin2017video}
Xiaojie Jin, Xin Li, Huaxin Xiao, Xiaohui Shen, Zhe Lin, Jimei Yang, Yunpeng
  Chen, Jian Dong, Luoqi Liu, Zequn Jie, et~al.
\newblock Video scene parsing with predictive feature learning.
\newblock In {\em Proceedings of the IEEE International Conference on Computer
  Vision}, pages 5580--5588, 2017.

\bibitem{khoreva2017lucid}
Anna Khoreva, Rodrigo Benenson, Eddy Ilg, Thomas Brox, and Bernt Schiele.
\newblock Lucid data dreaming for object tracking.
\newblock In {\em The DAVIS Challenge on Video Object Segmentation}, 2017.

\bibitem{kiani2013multi}
Hamed Kiani~Galoogahi, Terence Sim, and Simon Lucey.
\newblock Multi-channel correlation filters.
\newblock In {\em Proceedings of the IEEE international conference on computer
  vision}, pages 3072--3079, 2013.

\bibitem{kiani2015correlation}
Hamed Kiani~Galoogahi, Terence Sim, and Simon Lucey.
\newblock Correlation filters with limited boundaries.
\newblock In {\em Proceedings of the IEEE Conference on Computer Vision and
  Pattern Recognition}, pages 4630--4638, 2015.

\bibitem{koch2015siamese}
Gregory Koch, Richard Zemel, and Ruslan Salakhutdinov.
\newblock Siamese neural networks for one-shot image recognition.
\newblock In {\em ICML Deep Learning Workshop}, volume~2, 2015.

\bibitem{Kristan2018a}
Matej Kristan, Ales Leonardis, Jiri Matas, Michael Felsberg, Roman Pfugfelder,
  Luka~Cehovin Zajc, Tomas Vojir, Goutam Bhat, Alan Lukezic, Abdelrahman
  Eldesokey, Gustavo Fernandez, and et al.
\newblock The sixth visual object tracking vot2018 challenge results, 2018.

\bibitem{10.1007/978-3-319-48881-3_54}
Matej Kristan, Ji{\v{r}}i Matas, Ale{\v{s}} Leonardis, Michael Felsberg,
  Gustavo Fern{\'a}ndez, and et al.
\newblock The visual object tracking vot2016 challenge results.
\newblock In {\em Proceedings of the European Conference on Computer Vision
  Workshop}, pages 777--823. Springer International Publishing, 2016.

\bibitem{li2018siamrpn++}
Bo Li, Wei Wu, Qiang Wang, Fangyi Zhang, Junliang Xing, and Junjie Yan.
\newblock Siamrpn++: Evolution of siamese visual tracking with very deep
  networks.
\newblock {\em arXiv preprint arXiv:1812.11703}, 2018.

\bibitem{li2018high}
Bo Li, Junjie Yan, Wei Wu, Zheng Zhu, and Xiaolin Hu.
\newblock High performance visual tracking with siamese region proposal
  network.
\newblock In {\em Proceedings of the IEEE Conference on Computer Vision and
  Pattern Recognition}, pages 8971--8980, 2018.

\bibitem{li2014scale}
Yang Li and Jianke Zhu.
\newblock A scale adaptive kernel correlation filter tracker with feature
  integration.
\newblock In {\em European conference on computer vision}, pages 254--265.
  Springer, 2014.

\bibitem{liang2017single}
Yudong Liang, Ze Yang, Kai Zhang, Yihui He, Jinjun Wang, and Nanning Zheng.
\newblock Single image super-resolution via a lightweight residual
  convolutional neural network.
\newblock {\em arXiv preprint arXiv:1703.08173}, 2017.

\bibitem{luc2017predicting}
Pauline Luc, Natalia Neverova, Camille Couprie, Jakob Verbeek, and Yann LeCun.
\newblock Predicting deeper into the future of semantic segmentation.
\newblock In {\em Proceedings of the IEEE International Conference on Computer
  Vision}, pages 648--657, 2017.

\bibitem{ma2015long}
Chao Ma, Xiaokang Yang, Chongyang Zhang, and Ming-Hsuan Yang.
\newblock Long-term correlation tracking.
\newblock In {\em Proceedings of the IEEE conference on computer vision and
  pattern recognition}, pages 5388--5396, 2015.

\bibitem{marki2016bilateral}
Nicolas M{\"a}rki, Federico Perazzi, Oliver Wang, and Alexander
  Sorkine-Hornung.
\newblock Bilateral space video segmentation.
\newblock In {\em Proceedings of the IEEE Conference on Computer Vision and
  Pattern Recognition}, pages 743--751, 2016.

\bibitem{newton1999principia}
Isaac Newton.
\newblock {\em The Principia: mathematical principles of natural philosophy}.
\newblock Univ of California Press, 1999.

\bibitem{nijhawan1994motion}
Romi Nijhawan.
\newblock Motion extrapolation in catching.
\newblock {\em Nature}, 1994.

\bibitem{nilsson2018semantic}
David Nilsson and Cristian Sminchisescu.
\newblock Semantic video segmentation by gated recurrent flow propagation.
\newblock In {\em Proceedings of the IEEE Conference on Computer Vision and
  Pattern Recognition}, pages 6819--6828, 2018.

\bibitem{perazzi2017learning}
Federico Perazzi, Anna Khoreva, Rodrigo Benenson, Bernt Schiele, and Alexander
  Sorkine-Hornung.
\newblock Learning video object segmentation from static images.
\newblock In {\em Proceedings of the IEEE Conference on Computer Vision and
  Pattern Recognition}, pages 2663--2672, 2017.

\bibitem{perazzi2016benchmark}
Federico Perazzi, Jordi Pont-Tuset, Brian McWilliams, Luc Van~Gool, Markus
  Gross, and Alexander Sorkine-Hornung.
\newblock A benchmark dataset and evaluation methodology for video object
  segmentation.
\newblock In {\em Proceedings of the IEEE Conference on Computer Vision and
  Pattern Recognition}, pages 724--732, 2016.

\bibitem{pinheiro2015learning}
Pedro~O Pinheiro, Ronan Collobert, and Piotr Doll{\'a}r.
\newblock Learning to segment object candidates.
\newblock In {\em Advances in Neural Information Processing Systems}, pages
  1990--1998, 2015.

\bibitem{pinheiro2016learning}
Pedro~O Pinheiro, Tsung-Yi Lin, Ronan Collobert, and Piotr Doll{\'a}r.
\newblock Learning to refine object segments.
\newblock In {\em European Conference on Computer Vision}, pages 75--91.
  Springer, 2016.

\bibitem{pont20172017}
Jordi Pont-Tuset, Federico Perazzi, Sergi Caelles, Pablo Arbel{\'a}ez, Alex
  Sorkine-Hornung, and Luc Van~Gool.
\newblock The 2017 davis challenge on video object segmentation.
\newblock {\em arXiv preprint arXiv:1704.00675}, 2017.

\bibitem{ramanan2007tracking}
Deva Ramanan, David~A Forsyth, and Andrew Zisserman.
\newblock Tracking people by learning their appearance.
\newblock {\em IEEE transactions on pattern analysis and machine intelligence},
  29(1):65--81, 2007.

\bibitem{ren2015faster}
Shaoqing Ren, Kaiming He, Ross Girshick, and Jian Sun.
\newblock Faster r-cnn: Towards real-time object detection with region proposal
  networks.
\newblock In {\em Advances in neural information processing systems}, pages
  91--99, 2015.

\bibitem{smeets1998visuomotor}
Jeroen~BJ Smeets, Eli Brenner, and Marc~HE de Lussanet.
\newblock Visuomotor delays when hitting running spiders.
\newblock In {\em EWEP 5-Advances in perception-action coupling}, pages 36--40.
  {\'E}ditions EDK, Paris, 1998.

\bibitem{sun2018pwc}
Deqing Sun, Xiaodong Yang, Ming-Yu Liu, and Jan Kautz.
\newblock Pwc-net: Cnns for optical flow using pyramid, warping, and cost
  volume.
\newblock In {\em Proceedings of the IEEE Conference on Computer Vision and
  Pattern Recognition}, pages 8934--8943, 2018.

\bibitem{tao2016siamese}
Ran Tao, Efstratios Gavves, and Arnold~WM Smeulders.
\newblock Siamese instance search for tracking.
\newblock In {\em Proceedings of the IEEE conference on computer vision and
  pattern recognition}, pages 1420--1429, 2016.

\bibitem{tsai2016video}
Yi-Hsuan Tsai, Ming-Hsuan Yang, and Michael~J Black.
\newblock Video segmentation via object flow.
\newblock In {\em Proceedings of the IEEE conference on computer vision and
  pattern recognition}, pages 3899--3908, 2016.

\bibitem{valmadre2017end}
Jack Valmadre, Luca Bertinetto, Jo{\~a}o Henriques, Andrea Vedaldi, and
  Philip~HS Torr.
\newblock End-to-end representation learning for correlation filter based
  tracking.
\newblock In {\em Computer Vision and Pattern Recognition (CVPR), 2017 IEEE
  Conference on}, pages 5000--5008. IEEE, 2017.

\bibitem{van2018motion}
Elle van Heusden, Martin Rolfs, Patrick Cavanagh, and Hinze Hogendoorn.
\newblock Motion extrapolation for eye movements predicts perceived
  motion-induced position shifts.
\newblock {\em Journal of Neuroscience}, 38(38):8243--8250, 2018.

\bibitem{voigtlaender2017online}
Paul Voigtlaender and Bastian Leibe.
\newblock Online adaptation of convolutional neural networks for video object
  segmentation.
\newblock {\em arXiv preprint arXiv:1706.09364}, 2017.

\bibitem{von1982eye}
Claes Von~Hofsten.
\newblock Eye--hand coordination in the newborn.
\newblock {\em Developmental psychology}, 18(3):450, 1982.

\bibitem{walker2017pose}
Jacob Walker, Kenneth Marino, Abhinav Gupta, and Martial Hebert.
\newblock The pose knows: Video forecasting by generating pose futures.
\newblock In {\em Proceedings of the IEEE International Conference on Computer
  Vision}, pages 3332--3341, 2017.

\bibitem{6751553}
H. {Wang} and C. {Schmid}.
\newblock Action recognition with improved trajectories.
\newblock In {\em 2013 IEEE International Conference on Computer Vision}, pages
  3551--3558, Dec 2013.

\bibitem{wang2017prioritization}
Jianren Wang, Long Qian, Ehsan Azimi, and Peter Kazanzides.
\newblock Prioritization and static error compensation for multi-camera
  collaborative tracking in augmented reality.
\newblock In {\em 2017 IEEE Virtual Reality (VR)}, pages 335--336. IEEE, 2017.

\bibitem{wang2018vertical}
Jianren Wang, Junkai Xu, and Peter~B Shull.
\newblock Vertical jump height estimation algorithm based on takeoff and
  landing identification via foot-worn inertial sensing.
\newblock {\em Journal of biomechanical engineering}, 140(3):034502, 2018.

\bibitem{wang2018fast}
Qiang Wang, Li Zhang, Luca Bertinetto, Weiming Hu, and Philip~HS Torr.
\newblock Fast online object tracking and segmentation: A unifying approach.
\newblock {\em arXiv preprint arXiv:1812.05050}, 2018.

\bibitem{willems2008efficient}
Geert Willems, Tinne Tuytelaars, and Luc Van~Gool.
\newblock An efficient dense and scale-invariant spatio-temporal interest point
  detector.
\newblock In {\em European conference on computer vision}, pages 650--663.
  Springer, 2008.

\bibitem{wu2015object}
Yi Wu, Jongwoo Lim, and Ming-Hsuan Yang.
\newblock Object tracking benchmark.
\newblock {\em IEEE Transactions on Pattern Analysis and Machine Intelligence},
  37(9):1834--1848, 2015.

\bibitem{xia2017validation}
Haisheng Xia, Junkai Xu, Jianren Wang, Michael~A Hunt, and Peter~B Shull.
\newblock Validation of a smart shoe for estimating foot progression angle
  during walking gait.
\newblock {\em Journal of biomechanics}, 61:193--198, 2017.

\bibitem{zhu2019feature}
Chenchen Zhu, Yihui He, and Marios Savvides.
\newblock Feature selective anchor-free module for single-shot object
  detection.
\newblock {\em arXiv preprint arXiv:1903.00621}, 2019.

\bibitem{zhu2018distractor}
Zheng Zhu, Qiang Wang, Bo Li, Wei Wu, Junjie Yan, and Weiming Hu.
\newblock Distractor-aware siamese networks for visual object tracking.
\newblock In {\em European Conference on Computer Vision}, pages 103--119.
  Springer, 2018.

\end{thebibliography}
}

\end{document}